# Exploring the relationship between response time sequence in scale answering process and severity of insomnia: a machine learning approach

Zhao Su[1*], Rongxun Liu[2,3*], Keyin Zhou[2], Xinru Wei[1], Ning Wang[2,4], Zexin Lin[2], Yuanchen Xie[5], Jie Wang[1], Fei Wang[2#], Shenzhong Zhang[1#], Xizhe Zhang[1#]

[1]School of Biomedical Engineering and Informatics, Nanjing Medical University, Nanjing, Jiangsu, China.
[2]Early Intervention Unit, Department of Psychiatry, Affiliated Nanjing Brain Hospital, Nanjing Medical University, Nanjing, Jiangsu, China.
[3]School of Psychology, Xinxiang Medical University, Xinxiang, Henan, China
[4]School of Public Health, Xinxiang Medical University, Xinxiang, Henan, China
[5]The Fourth School of Clinical Medicine, Nanjing Medical University, Nanjing, Jiangsu, China.
*The two authors contributed equally to this work: Zhao Su, Rongxun Liu
#Corresponding authors: Fei Wang fei.wang@yale.edu, Shenzhong Zhang zsz@njmu.edu.cn, Xizhe Zhang zhangxizhe@njmu.edu.cn.

## Abstract

**Objectives:** The study aims to investigate the relationship between insomnia and response time. Additionally, it aims to develop a machine learning model to predict the presence of insomnia in participants using response time data.
**Methods:** A mobile application was designed to administer scale tests and collect response time data from 2729 participants. The relationship between symptom severity and response time was explored, and a machine learning model was developed to predict the presence of insomnia.
**Results:** The result revealed a statistically significant difference (p<.001) in the total response time between participants with or without insomnia symptoms. A correlation was observed between the severity of specific insomnia aspects and response times at the individual questions level. The machine learning model demonstrated a high predictive accuracy of 0.743 in predicting insomnia symptoms based on response time data.
**Conclusions:** These findings highlight the potential utility of response time data to evaluate cognitive and psychological measures, demonstrating the effectiveness of using response time as a diagnostic tool in the assessment of insomnia.

## Introduction

Psychological self-assessment scales are widely used to evaluate various mental health factors, including insomnia. However, these scales often come with the subjective bias of the participants, leading to potential inconsistencies in severity and symptom expression [1,2]. Especially for self-assessment scales, the reports of people on their own thoughts, feelings, and behaviors often result in subjective bias [3]. For instance, the Insomnia Severity Index (ISI) scale, employed in this study, may yield divergent assessments of insomnia severity and symptoms even when two individuals report identical scores. With the advent of information technology, participants can conveniently answer assessment scales on their personal device [4]. At the same time, their behavioral data during the answering process can also be collected [5]. This behavioral data can offer an objective perspective on the participant's state during the scale completion, and could significantly assist in the interpretation of scale results [6].

Response Time (RT), which represents the duration taken by a participant to respond to a stimulus or complete a task, is a meaningful metric commonly used in various experiments in cognitive psychology [7]. In cognitive psychology research, RT functions as a dependent variable, influenced by manipulations of independent variables like stimulus exposure duration [8]. It shares a relationship with response accuracy (another primary dependent variable), because participants can often trade off speed for increased accuracy, or conversely, trade off accuracy for increased speed [9]. However, it's important to highlight that RT and accuracy often serve divergent objectives in these studies. The RT used in this study deviates from that in traditional psychology experiments. Rather than measuring the RT to a singular event, it records the timing of a sequence of responses, specifically the process of answering a scale. In the process of participants answering the scale, the data is unobtrusively collected, implying it can be used as an objective biomarker for identifying insomnia symptoms. For example, RT can mirror complex thought processes, thereby acting as a reflection of an individual's cognitive abilities. It is noteworthy that insomnia, a prevalent sleep disorder, can induce daytime fatigue and decrease cognitive functioning [10], which may consequently influence response times. These effects might change how quickly a person responds to things. Given this connection, RT measures could serve as a complementary metric to the Insomnia Severity Index (ISI) results, potentially enhancing the scale's accuracy and reliability. Furthermore, RT can independently indicate insomnia severity, thereby contributing another dimension of objectivity to the assessment process.

An expanding body of research recognizes the utility of Response Time (RT) data. For instance, researchers have employed RT as a metric for understanding human information processing [9]. Previous studies have examined the correlation between RT and cognitive load, uncovering that extended RTs often signify more complex cognitive processes [11]. Within psychological assessments, research suggests that RTs can signify indecision or emotional conflict, thus offering a deeper understanding of individual responses [11,12]. Moreover, an emergent research trend investigates the relationship between mobile phone usage behavior and mental health, including insomnia symptoms [13,14].

In this study, we gathered Response Time (RT) data for each question during the participants' response process. The ensuing time interval sequences often encapsulate a plethora of information. Prior research has harnessed RT data to filter out invalid responses [15–18], underscoring the value of RT as a metric for assessing participants' cognitive and psychological states during different evaluations. This data can contribute to a more precise understanding of an individual's mental state and self-perception, ultimately enhancing the accuracy of diagnoses and treatment plans. Machine Learning models are particularly apt at discerning underlying patterns between RT sequences and final scale outcomes. Despite this, few studies have exploited RT data for predicting final scale results.

The primary objective of this study is to elucidate the relationship between Response Time (RT) and the results of Insomnia Severity Index (ISI) scale assessments. To probe this relationship, we employ regression analysis to evaluate how RT correlates with the responses on the ISI scale. We hypothesize that certain scale items may trigger longer or shorter RTs. To validate our findings, we develop machine learning models capable of predicting the presence of insomnia using RT data. Additionally, we conduct a comprehensive analysis of the most impactful features in these predictive models to gain insights into their relative importance and interactions that influence prediction outcomes. Our research explores the utility of RT data as a novel tool for evaluating cognitive and psychological measures, particularly in the context of sleep disorders. Ultimately, this study underscores the potential of RT data in augmenting cognitive and psychological assessment practices, thereby contributing to more accurate diagnoses and effective treatment strategies for sleep disorders.

## Participants and Methods

### Data Acquisition

The methodology of this study revolves around a comprehensive psychological screening program designed specifically for new enrollees at Nanjing Medical University. All the participants are newly enrolled first-year university students. A substantial sample size was employed, involving a total of 2729 participants. The average age in this group is 19, with a standard deviation of 0.93. All data collection took place within the university campus. Finally, 14,707 interval values were generated and analyzed. The participants interacted with the screening scale through a dedicated application installed on their personal cell phones.

For data acquisition, we used the Insomnia Severity Index (ISI) scale (Chinese version) to gauge the severity of insomnia symptoms among our participants [19]. The ISI is composed of seven questions, each having five response options ranging from 0 to 4 points, with higher scores denoting increased severity of insomnia symptoms.

The total score is computed by summing the scores from all seven questions, and participants garnering a total score of 7 or higher were deemed to exhibit insomnia symptoms.

## Response Time

The scale was administered via a mobile platform. When a participant moves into a new question, a timestamp is recorded, and another timestamp is taken when the participant selects an answer to this question. The difference between these two timestamps equates to the value of the RT interval. Since the data collection process is conducted online through the Internet, we recognize that following the timestamp marking the start, static resources like the program's UI are loaded, which could slightly affect the RT values due to network latency. The data collected in this study is all from local university students who answered the scale on campus, minimizing impact of network latency on the data.

The procedure generated a sequence of 7 RTs for each participant. Additionally, the system recorded the chosen response for each question, facilitating the calculation of the total score.

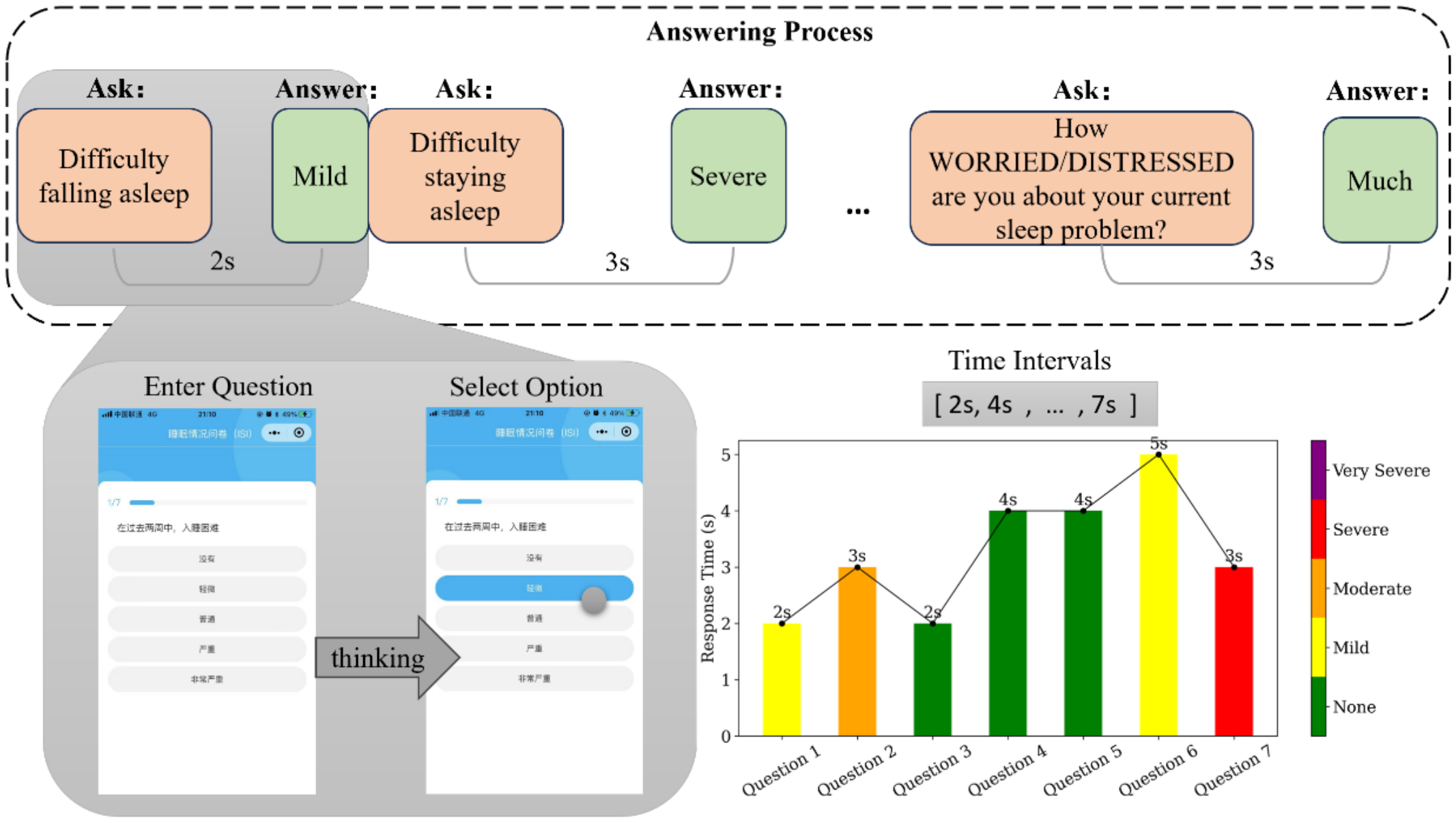

Figure 1. Response time intervals generating in participant answer processing

## Data Processing

*Data Cleaning:* In this study, the data collection process was dependent on participant responses, which can introduce a range of issues such as network interruptions or dropout cases. The resulting missing values from these instances compromise the quality of the data. To address this, we deployed data cleaning methods that excluded participants with such data irregularities from the study. Additionally, prolonged pauses in a participant's response, exceeding 60 seconds, can signify an outlier in the data, and are also excluded to maintain the desired level of reliability and validity. These cleaning measures ensure that the data used in the analysis is devoid of any significant irregularities, reducing the risk of biased or inconsistent results.

*Exclusion of Careless Responses:* Further, we also took measures to exclude careless responses from the dataset. The methods used to identify such responses involve detecting RTs that were unusually low or highly variable. Specifically, if a participant's average RT is below 1.5 seconds or the variance of their RTs exceed 6 seconds, that participant's data will be excluded from the analysis to avoid confounding results[20]. By doing so, we can ensure that only data from participants who demonstrated an adequate level of engagement and conscientiousness during the study are included for analysis.

## Statistics Analysis

In order to compare differences in RT between the two groups – individuals with insomnia symptoms and those without – we employed a non-parametric statistical test, the Mann-Whitney U test. This choice of this test was informed by the non-normal distribution of the data and the ordinal nature of the variables under investigation. This test was used to discern the outcome scores between two independent groups. A significant level of 0.01 was set a priori. Following this, we implemented a one-way Analysis of Variance (ANOVA) with an F-test to examine differences in RT across multiple groups. These groups were determined based on the five different options that participants could select during the task. To further explore, we applied quadratic regression analysis to explore the relationship between option selection (treated as an ordinal variable) and RT. We chose a quadratic model considering the potential nonlinearity between option selection and RT. All statistical analyses were performed using the Python and the statsmodels package [21].

## Feature Calculation

Our analysis aimed to derive deeper insights into the RT sequence data we collected. For this purpose, we calculated a range of statistical features that would help to describe the distribution of RT. Specifically, we computed the mean, variance, maximum, minimum, median, skewness, kurtosis, range, and coefficient of variation. These features are commonly used in exploratory data analysis and can provide a detailed overview of the data. We also compute features called 'freq_x', which denotes the frequency of a given value 'x'. For instance, 'freq_1' signifies the proportion of RT falling within the 1-2 s range in relation to the total RT sequence. Similarly, 'big_than_x' features was also calculated, which denote the frequency of values in the sequence that are greater than 'x'. We also used dimensionality reduction on the high-dimensional RT data. We used two techniques, PCA [22] and t-SNE [23] to identify hidden patterns and to visualize the data in a lower-dimensional space. By applying these methods, we were able to achieve a more detailed understanding of the RT data. Further, the information extracted from the RT data was used as input features for subsequent machine learning models.

## Prediction Model

Results were reported for both raw data and extracted features. Results were reported for both approaches. Utilizing extracted features for model building enhanced interpretability, which is particularly important in the medical field. The participants' scores were evaluated from seven questions and summed up to constitute the total score. Participants who obtained a score of 7 or more were determined to exhibit symptoms of insomnia and were accordingly labeled as '1'. All other participants were assigned the label '0'.

To avoid the effects of label imbalance, the entire dataset was down sampled 10 times (total sample size: 332, with 166 samples for each label). 10-fold cross-validation was employed to validate the models on the dataset. Five algorithms were compared in this study, including logistic regression, decision trees, support vector machines, K-nearest neighbors and neural networks (Multi-layer Perceptron). For hyperparameters tuning, we used Optuna, an automatic optimization tool based on the Bayesian optimization algorithm [24]. The model construction experiment utilized scikit-learn [25], imblearn [26], and mlxtend [27]. The hyperparameters for each model, as determined by tuning, are shown in Supplement Table A. A detailed overview of the modeling steps is presented in Figure 2.

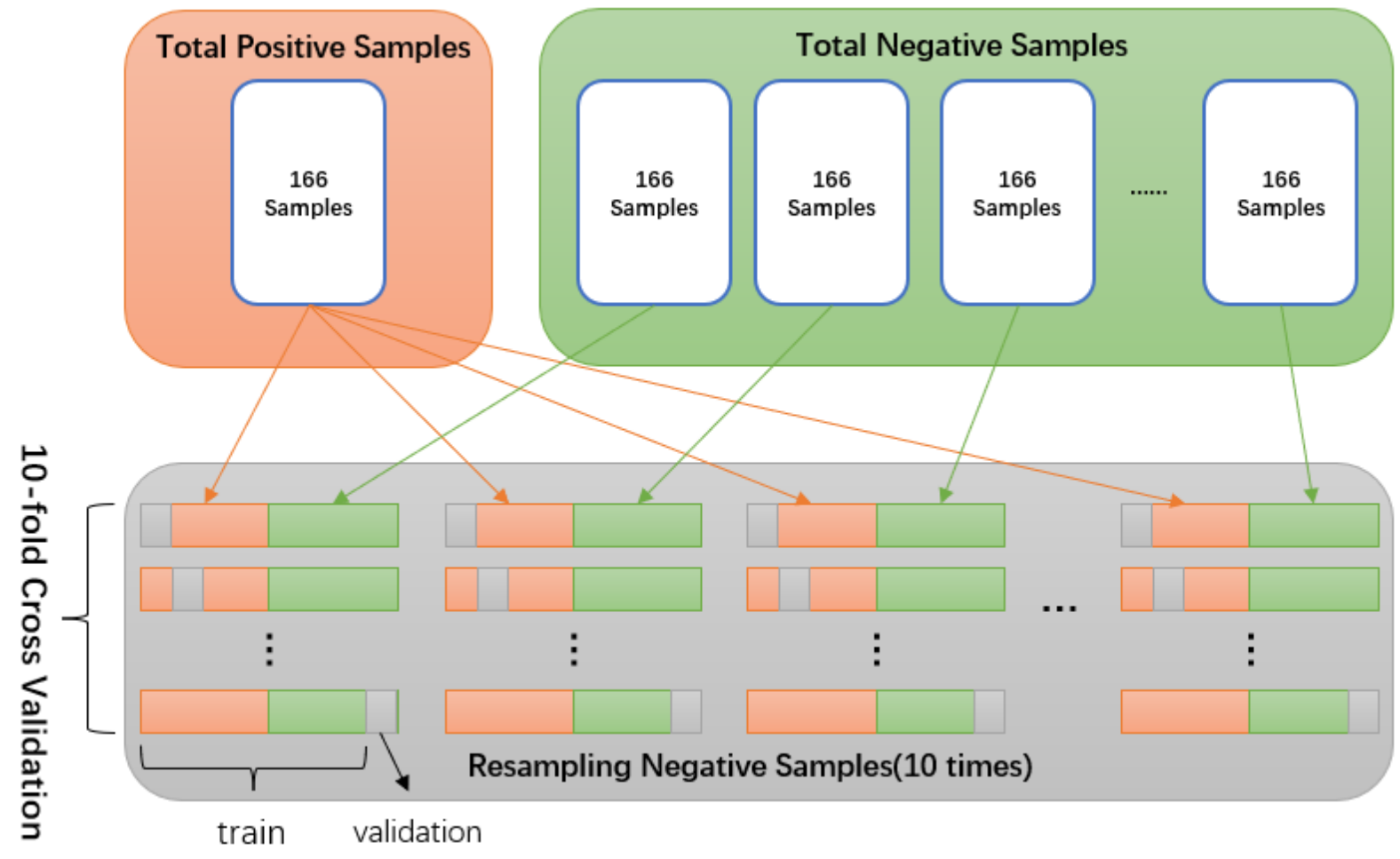

Figure 2. Data downsampling and 10-fold cross-validation

## Feature Selection

To select the most predictive feature subset from the original feature set, we employed the sequential feature selection algorithm for feature selection [28]. This algorithm iteratively selects and refines features based on the current feature subset, gradually reducing the size of the feature set and ultimately obtaining the most predictive feature subset. We utilized logistic regression as the estimator with a limitation of 10 features. The procedure was conducted using backward feature selection without implementing floating feature selection, and the model performance was assessed using the r2 metric, validated via 3-fold cross-validation.

## Results

### Demographic Statistics

A total of 2729 participants contributed data to this study. Out of the 2729 individuals recruited, 2101 met the inclusion criteria for the study, 166 were labeled as having insomnia, while 1935 were labeled as not having insomnia. Additional details regarding the study participants can be found in Table 2. Among the analyzed participants, 966 were men (86 displaying symptoms of insomnia), and 1135 were women (80 displaying symptoms of insomnia). The participants in the experiment are all first-year university students who have just enrolled, with ages ranging from 18 to 21 years old. Finally, 14,707 interval values were generated and analyzed. T-tests or F-tests were employed to examine whether the distribution of ISI scores was consistent across different demographic groups within our sample population. Further details are available in Table 1.

Table 1. Number of participants with different labels and compliance with exclusion criteria

| | **Total** | **Excluded** | **Included** |
|---|---|---|---|
| non-insomnia | 2548 | 613 | 1935 |
| insomnia | 181 | 15 | 166 |

Table 2. Demographics of the participants

| | Total(n=2101) | Insomnia(n=166) | F or t value | P value |
|---|---|---|---|---|
| **Gender** | | | 1.574 | 0.116 |
| male | 966(45.978%) | 86(51.807%) | | |
| female | 1135(54.022%) | 80(48.193%) | | |
| **History psychological illness** | | | -5.763 | <0.001 |
| no | 2078(98.905%) | 157(94.578%) | | |
| yes | 23(1.095%) | 9(5.422%) | | |
| **History physical illness** | | | -4.889 | <0.001 |
| no | 2038(97.001%) | 152(91.566%) | | |
| yes | 63(2.999%) | 14(8.434%) | | |
| **Smoke** | | | 12.389 | <0.001 |
| never | 2083(99.143%) | 160(96.386%) | | |
| occasional | 12(0.571%) | 2(1.205%) | | |
| frequent | 4(0.19%) | 3(1.807%) | | |
| former | 2(0.095%) | 1(0.602%) | | |
| **Drink** | | | 15.757 | <0.001 |
| never | 1607(76.487%) | 108(65.06%) | | |
| occasional | 470(22.37%) | 55(33.133%) | | |
| frequent | 3(0.143%) | 1(0.602%) | | |
| former | 21(1.0%) | 2(1.205%) | | |

## The Relationship Between Score and Response Time

We utilized the Mann-Whitney U test to assess the differences in total RT between individuals with and without insomnia symptoms. The test revealed a significant difference in the distribution between these two independent groups ($U = 93856.5$, $p < .001$). Specifically, the median score for the insomnia group (Median = 28, IQR = 11.75) was statistically significantly higher compared to the non-insomnia group (Median = 22, IQR = 12).

Each question had five options, corresponding to scores from 0 to 4, where higher scores indicating more severe conditions. First, we conducted an analysis of variance to compare RTs among the different options for each of the seven questions. All questions showed a statistically significant effect, $p < .001$. Questions 3 and 4 showed the highest effect sizes with F-values of 164.358 and 168.108 respectively. Questions 1, 2, 5, and 7 also showed significant effects with F-values ranging from 40.781 to 83.792. Question 6 had a significant effect with a smaller F-value of 4.591 but still significant at $p=.001$. All results are summarized in Supplement Table B.

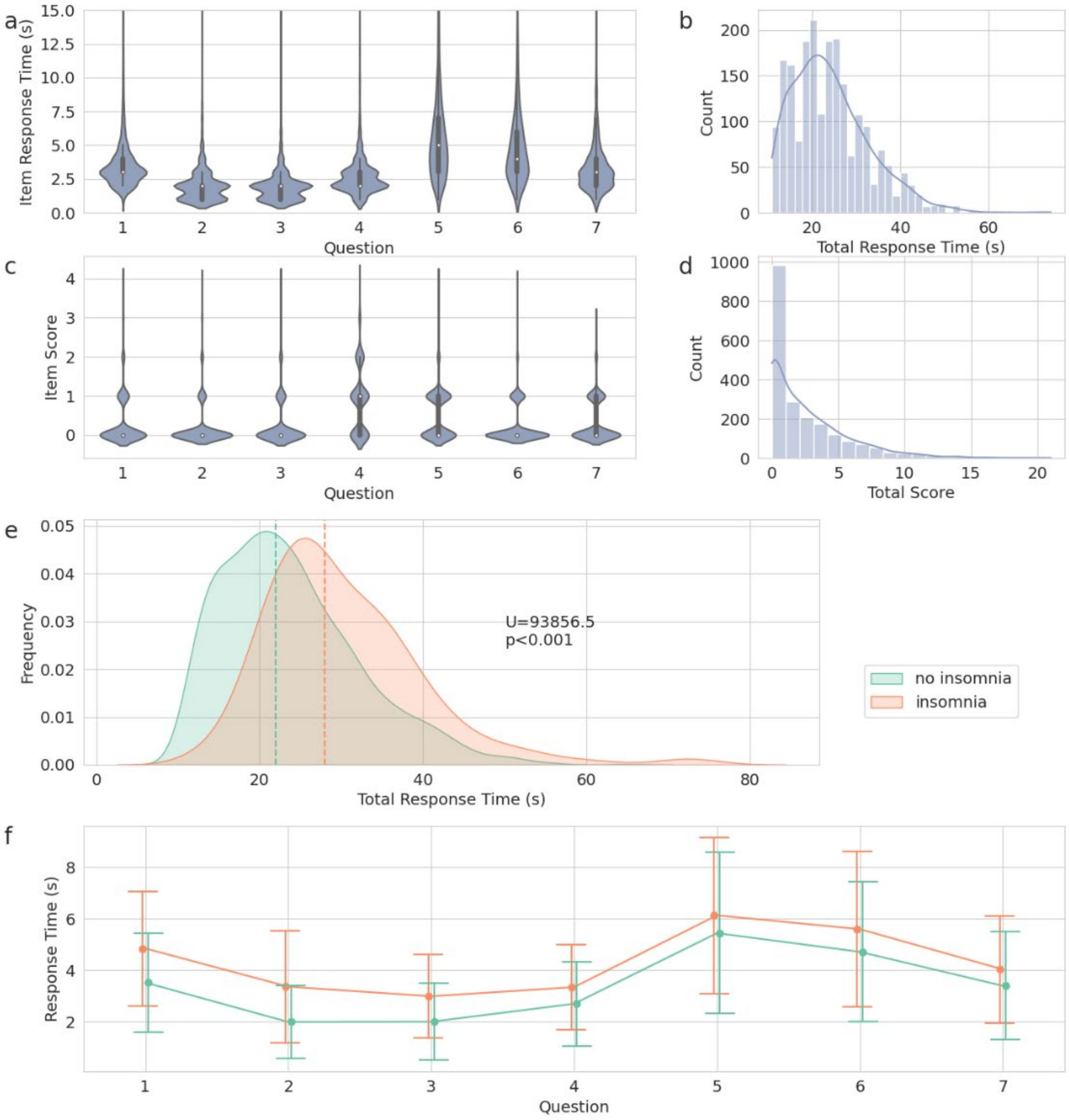

Figure 3. Comparative analysis of response times and scores on the Insomnia Severity Index (ISI) for groups with and without symptoms of insomnia. **a)** Violin plot representing the distribution of RTs for each of the seven questions on the ISI scale. Each "violin" represents a different question, with the width of the plot at a given point indicating the density of response times at that value. **b)** Histogram showing the overall distribution of RTs for finishing the total ISI scale. The x-axis represents the total time taken to answer all seven questions, while the y-axis represents the frequency of each RT. **c)** Violin plot representing the distribution of scores for each of the seven questions on the ISI scale. Similar to a), each "violin" represents a different question, but here, the width of the plot at a given point indicates the density of scores at that value. **d)** Histogram showing the overall distribution of total scores on the ISI scale. The x-axis represents the score of ISI, and the y-axis indicates the frequency of each total score. **e)** Distribution plot comparing the ISI score (y-axis) versus RT distribution (x-axis) for two groups: those with symptoms of insomnia (marked in orange) and those without symptoms (marked in a green). The median values for each group are highlighted. **f)** Showing the differences in response times for individual questions between the two groups

(with and without symptoms of insomnia). Each pair of bars represents a different question from the ISI scale, with the point of the bar indicating the average RT.

We investigated the relationship between the variables single question score (which stands by x) and single question RT (which stands by y) by fitting a quadratic regression model to the data. The model was specified as follows: $y = \beta0 + \beta1 * x + \beta2 * x^2 + \varepsilon$. Our sample consisted of 2,101 observations. Focusing on the first item as an example, the overall model was significant ($F = 74.76$, $p < .001$) and accounted for approximately 6.7% of the variance in the dependent variable y ($R^2 = 0.067$, adjusted $R^2 = 0.066$). The linear term (x) was statistically significant with a coefficient of 1.179 ($t = 7.886$, $p < .001$, 95% CI [0.886, 1.472]), indicating a positive relationship between x and y. The quadratic term ($x^2$) was also statistically significant with a coefficient of -0.183 ($t = -2.826$, $p = .005$, 95% CI [-0.310, -0.056]) indicating a U-shaped relationship between score and RT, in line with our hypothesis. Detailed results for each item are shown in Figure 4.

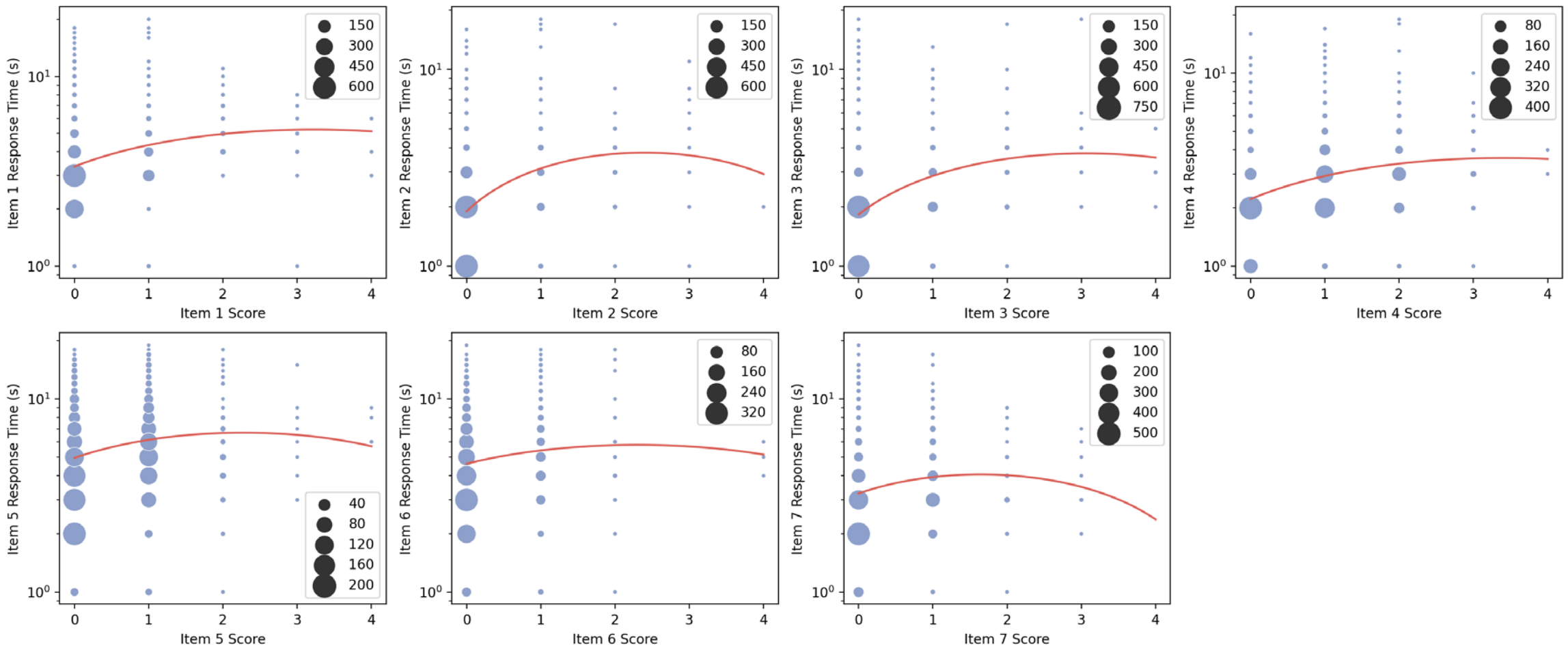

Figure 4. Results of regression analysis for item score and item response time. The figure presents the relationship between the score of the participants for questions 1-7 of the ISI scale and the RT. Each of the seven models was significant, explaining 6.7%, 10.7%, 11.1%, 8%, 4%, 1.5%, and 2.3% of the variance in the dependent variable y, respectively. The coefficients of the linear term (x) were 1.17, 1.56, 1.25, 0.82, 1.53, 1.01, and 0.99 respectively; while the coefficients for the quadratic term ($x^2$) were -0.18, -0.32, -0.20, -0.12, -0.33, -0.22, and -0.3 respectively.

## Statistical Analysis of Features

In the initial step, we identified features with a correlation coefficient above 0.8, resulting in 27 key features listed in Supplement Table C. A subsequent examination of interrelationships among these features revealed substantial correlations. To compare the groups with and without insomnia, we applied T-tests to each of the 27 features. The results revealed significant differences between the two groups on several features. For example, individuals with insomnia exhibited higher levels of mean of RT values and lower levels of coefficient of variation (cv) of RT values compared to those without insomnia. Finally, we performed correlation analyses between each of the 27 features and the severity of insomnia (the scores on the ISI scale). The results indicate that several features were significantly correlated with the severity of insomnia which stand by ISI scores. All the results shown in Supplement Table C.

## Classification Results

Figure 5 present performance evaluations for the models employed in our study: logistic regression, decision tree, support vector machine (SVM), k-nearest neighbors (KNN), and multi-layer perceptron (MLP). The evaluation was done using Receiver Operating Characteristic (ROC) curves for two categories of models: feature-based and raw data-based. In the feature-based category, the logistic regression and the multi-layer perceptron (MLP) models demonstrated the most proficient performance. The logistic regression model achieved an Area Under the ROC Curve (AUROC) of 0.811, suggesting a high true positive rate relative to the false positive rate, thus underscoring the model's excellent ability in distinguishing between the classes. The MLP model also presented a commendable performance with an AUROC of 0.8. To the raw data-based models, the performances were similar across the various models. Notably, the MLP model came out on top, with an AUROC of 0.809, indicating an excellent balance between sensitivity and specificity. This underscores the effectiveness of MLP in understanding the intricacies and non-linearities of raw data.

In feature prediction (Supplement Table D), logistic regression emerged as the most effective model, offering superior performance in accuracy (0.743), precision (0.744), and F1 score (0.713). It also demonstrated competitive recall (0.695). These results suggest that logistic regression excelled at accurate classification and balancing precision and recall.

However, with raw data prediction (Supplement Table E), the MLP model achieved the highest accuracy (0.735) and precision (0.733). Logistic regression maintained high precision (0.733), while the decision tree model excelled in recall (0.758). Interestingly, the decision tree model showed substantial improvement in recall when employing raw data (0.758) compared to feature prediction data (0.705). This suggests that the decision tree model was particularly effective at identifying true positives when raw data was utilized.

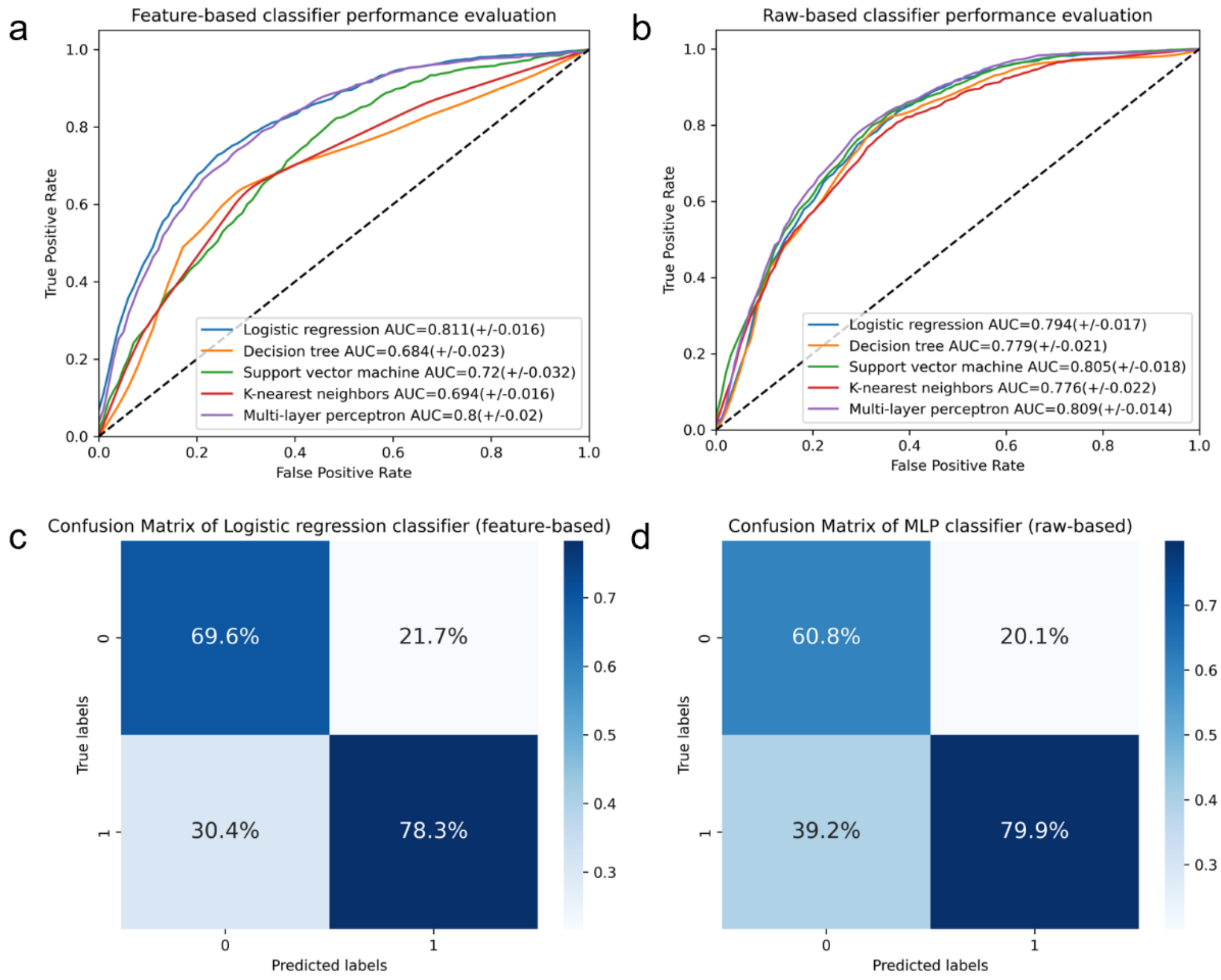

Figure 5. Performance comparison of various machine learning models on processed and raw response time interval data. **a)** ROC curves of Logistic Regression, Decision Tree, Support Vector Machine, K-nearest Neighbors, and Multi-layer Perceptron models applied to features derived from Response Time Interval data. **b)** ROC curves of the same models applied directly to raw Response Time Interval data. **c)** Confusion matrix of the best performing model (Logistic Regression) trained on the feature data. **d)** Confusion matrix of the best performing model (Multi-layer Perceptron) trained on raw data. All plots were obtained using 10-fold cross-validation, with the process repeated 10 times for resampling purposes.

## Feature Importance

In this study, we utilized Shapley Additive Explanations (SHAP) values to evaluate the contribution of each feature in the model [29]. Figure 6 illustrates the contribution rates of all the features mentioned in this paper using a logistic regression model. The top five features with the highest contribution rates are mean, pca_3, RT6_trans, tsne_3 and freq_2. If feature selection was conducted before model construction, multi features would have been excluded in the feature selection stage. These findings suggest that several features in the RT sequence are effective representations for insomnia symptoms.

The mean of RT emerged as the most important feature, indicating that the overall average speed of a participant's response plays a crucial role in the predictive model. Notably, the third principal component, pca_3, accounted for approximately 12.2% of the total variance in the data, representing a significant portion of the information in the dataset. Specifically, The pca_3 feature is primarily composed of the first, second, and third original features with corresponding loadings of 0.7105, 0.3003, and 0.3046, respectively. These loadings suggest that these features are the most relevant for pca_3, given their relatively high positive correlation. Interestingly, the fifth and sixth original features appear to contribute negatively to pca_3, with loadings of -0.1752 and -0.4141, respectively. Additionally, regarding the sixth question on " How WORRIED/DISTRESSED are you about your current sleep problem? ", this implies that this specific question on the ISI scale holds significant predictive value. The particular importance of this feature suggests a profound link between insomnia severity and the variable our model predicts. The proportion of RT sequence within the 1-3 seconds range, also emerged as highly significant. This result underscores the relevance of the timely reaction category of 1-3 seconds, suggesting that responses within this time frame carry substantial predictive weight. Lastly, the model identified the 25th percentile of RT as one of the most influential features, signifying that not only average RTs but also lower quartile RTs (i.e., faster responses) play a crucial role in outcome prediction.

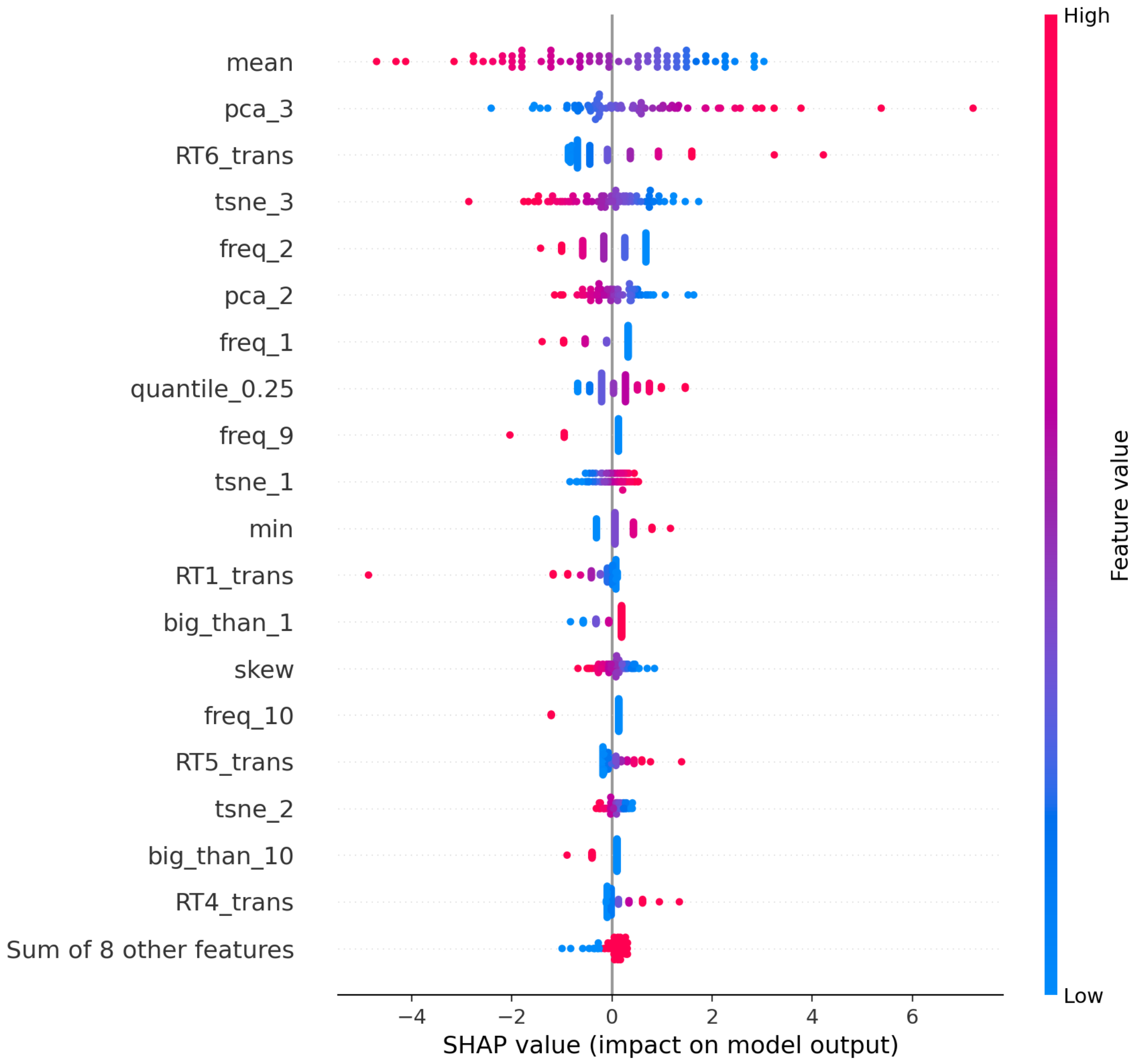

Figure 6. SHAP values of top features of best model

## Discussion

The key finding of this study lies in the utility of user behavior during the process of responding to a scale, specifically, the response time (RT), as an effective predictor of Insomnia Severity Index (ISI) scale results. The machine learning classifier model constructed achieved a remarkable accuracy rate of up to 74% and a recall rate of 75%. Previous studies related to this work mainly fall into two categories: the first assesses individuals' psychological or cognitive states based on prolonged mobile phone use behaviors, while the second focuses on using RT to screen for invalid scale responses - the same data type employed in this study. To date, no research has leveraged user behavior patterns during scale completion to explain or support scale outcomes. Our preliminary statistical test and regression analysis have identified a quadratic relationship pattern between RT and option choices, consistent with previous literature[6]. By exploring the interpretability of machine learning models, we found that features related to the length of RT, such as mean, tend to have high weights. This observation corroborates prior studies suggesting insomnia often results in a decline in daytime cognitive abilities[10].

In machine learning experiments, we observed that the feature-based model achieved a highest accuracy of 74% (via Logistic Regression model), outperforming the model based on raw response time interval series (via Neural Network model), which reached an accuracy of 73%. This observation suggests that, despite the superior fitting capacity of neural networks and their ability to effectively represent non-linear relationships, feature engineering sometimes can yield unexpected results, and additionally provides enhanced interpretability. The computation of statistical features, distribution features, and manually designed features of the response time interval series revealed a linear relationship with the labels. This made the machine learning models simpler and more effective, often leading to greater reliability in practical applications.

Our study has important implications for the design of psychological assessment tools and diagnostic criteria for insomnia. By identifying the most effective RT representations for insomnia symptoms, our findings can pave the way for creating innovative, more sensitive, and reliable assessment tools. Additionally, our study highlights the importance of considering RT data in the scale assessment, as RT can serve as an objective indicator of the condition's severity. Furthermore, the use of SHAP values can provide a more comprehensive evaluation of feature contributions to the model, which may further improve the accuracy of future diagnostic criteria. From a medical perspective, our findings open a new avenue for distinguishing symptoms of insomnia, uniquely leveraging existing RT data without the requirement for further evaluations by healthcare professionals. In the context of psychological assessment scales, our results advocate for the use of response time as an instrumental guide in scale design. Methodologically, the collection of participant behavior data during scale responses provides a complementary viewpoint, to complement scale results. This approach aids in enhancing the precision of scale evaluations and lessening the effect of subjectivity, offering potential improvements to the current assessment systems.

There are some notable limitations to our approach that should be considered. Firstly, our reliance on self-reported data poses a constraint, as such data can be prone to bias and may not faithfully represent an individual's sleep patterns. Secondly, the limited size of our sample for different age means our findings may not extend to larger populations. Another potential limitation of our study is that our model relies solely on RT as a predictor of insomnia severity. Future research could benefit from including other factors, such as detailed behaviors when answering the scale. Despite these limitations, our study lays a foundational groundwork for future investigations, facilitating the development of our understanding of the relationship between RT and insomnia severity.

As psychological or cognitive-related disorders such as insomnia become increasingly prevalent, the development of advanced tools and algorithms can offer more dependable and objective measurements of sleep quality. Wearable devices, capable of tracking physiological measures such as heart rate variability and brain waves, can offer more accurate and comprehensive sleep data. Moreover, machine learning algorithms can be utilized on large datasets to discern patterns and predictors of psychological or cognitive disorders. Overall, the collection of objective behavioral data paves the way for more precise diagnosis, enabling personalized treatment approaches to become a possibility.

## Conclusions

In conclusion, our study underscores the potential utility of response time (RT) as an effective indicator of insomnia severity. The findings presented herein suggest that machine learning techniques can substantially contribute to a more efficient and straightforward interpretation of RT data, thereby bolstering and expanding the implications of ISI scale results. We anticipate that as technology continues to advance, more personalized and data-driven approaches to insomnia assessment and treatment will emerge. The integration of behavioral data and machine learning algorithms could pave the way for reliable and objective methods of monitoring psychological and behavioral aspects of insomnia.

## Supplement

Table A. Hyperparameters for each model

| **Model** | **Parameter Name** | **Value of raw-based model** | **Value of feature-based model** |
|---|---|---|---|
| **Logistic Regression** | C | 6.38 | 0.16 |
| | class_weight | None | None |
| | dual | False | False |
| | fit_intercept | True | True |
| | max_iter | 100 | 300 |
| | penalty | l2 | l2 |
| | solver | lbfgs | lbfgs |
| | tol | 0.0001 | 0.0001 |
| **Decision Tree** | criterion | gini | gini |
| | max_depth | 8 | 1 |
| | min_samples_leaf | 4 | 2 |
| | min_samples_split | 10 | 3 |
| | splitter | best | best |
| **Support Vector Machine** | C | 1.15 | 6.76 |
| | gamma | 0.01 | 0.03 |
| | kernel | rbf | rbf |
| | tol | 0.001 | 0.001 |
| **K-nearest Neighbor** | algorithm | kd_tree | ball_tree |
| | leaf_size | 27 | 32 |
| | metric | minkowski | minkowski |
| | n_neighbors | 10 | 3 |
| | p | 2 | 2 |
| | weights | distance | uniform |
| **Multi-layer Perceptron** | activation | logistic | identity |
| | alpha | 0.0021 | 0.0011 |
| | hidden_layer_sizes | 150 | 100 |
| | learning_rate | adaptive | adaptive |
| | learning_rate_init | 0.001 | 0.006 |
| | max_iter | 129 | 274 |
| | solver | adam | adam |
| | tol | 0.0001 | 0.0001 |

Table B. F-test for RT of different option

| **Question** | **f_statistic** | **p_value** |
|---|---|---|
| Question1 | 39.63 | <.001 |
| Question2 | 64.20 | <.001 |
| Question3 | 66.89 | <.001 |
| Question4 | 48.32 | <.001 |
| Question5 | 22.76 | <.001 |
| Question6 | 10.33 | .001 |
| Question7 | 16.72 | <.001 |

Table C. T-test and Correlational Analysis for Features

| | | **T-test** | | **Correlational analysis** | |
|---|---|---|---|---|---|
| | **mean** | **t-statistic** | **p-value** | **corr** | **p-value** |
| **mean** | 3.467+/-1.254 | -8.4 | <.001 | 0.348 | <.001 |
| **min** | 1.603+/-0.666 | -10.971 | <.001 | 0.431 | <.001 |
| **skew** | 0.870+/-0.844 | 1.612 | .109 | -0.012 | .597 |
| **kurt** | 0.771+/-2.399 | 0.866 | .388 | -0.006 | .77 |
| **cv** | 0.549+/-0.206 | 7.409 | <.001 | -0.172 | <.001 |
| **freq_1** | 0.939+/-1.126 | 17.26 | <.001 | -0.388 | <.001 |
| **big_than_1** | 6.061+/-1.126 | -17.26 | <.001 | 0.388 | <.001 |
| **freq_2** | 2.076+/-1.318 | 10.127 | <.001 | -0.308 | <.001 |
| **freq_7** | 0.229+/-0.487 | -3.15 | .002 | 0.123 | <.001 |
| **freq_8** | 0.128+/-0.366 | -2.842 | .005 | 0.109 | <.001 |
| **freq_9** | 0.097+/-0.319 | -1.304 | .194 | 0.069 | .001 |
| **freq_10** | 0.069+/-0.258 | 0.892 | .374 | 0.018 | .421 |
| **cum_freq_10** | 6.826+/-0.434 | 1.44 | .151 | -0.076 | <.001 |
| **big_than_10** | 0.174+/-0.434 | -1.44 | .151 | 0.076 | <.001 |
| **quantile_0.25** | 2.182+/-0.829 | -12.09 | <.001 | 0.479 | <.001 |
| **pca_2** | 0.061+/-2.317 | -1.566 | .119 | 0.032 | .148 |
| **pca_3** | 0.095+/-1.997 | -7.864 | <.001 | 0.252 | <.001 |
| **tsne_1** | 0.560+/-8.680 | -4.479 | <.001 | 0.184 | <.001 |
| **tsne_2** | 2.985+/-7.473 | 2.169 | .031 | 0.042 | .052 |
| **tsne_3** | -1.758+/-6.784 | 12.452 | <.001 | -0.328 | <.001 |
| **RT1_trans** | 4.432+/-22.156 | -3.232 | .001 | 0.116 | <.001 |
| **RT2_trans** | 2.623+/-15.529 | -3.034 | .003 | 0.155 | <.001 |
| **RT3_trans** | 2.780+/-16.186 | -2.033 | .043 | 0.088 | <.001 |
| **RT4_trans** | 3.552+/-17.743 | -2.137 | .034 | 0.075 | <.001 |
| **RT5_trans** | 12.645+/-34.594 | -1.297 | .196 | 0.089 | <.001 |
| **RT6_trans** | 9.938+/-28.483 | -2.225 | .027 | 0.124 | <.001 |
| **RT7_trans** | 5.710+/-21.439 | -1.791 | .075 | 0.061 | .005 |

Table D. Performance evaluation of different models using raw data prediction results

| | accuracy | precision | recall | f1 | auc |
|---|---|---|---|---|---|
| Logistic Regression | 0.743 (+/-0.016) | 0.744 (+/-0.025) | 0.695 (+/-0.014) | 0.713 (+/-0.016) | 0.811 (+/-0.016) |
| Decision Tree | 0.708 (+/-0.033) | 0.724 (+/-0.068) | 0.632 (+/-0.04) | 0.666 (+/-0.024) | 0.684 (+/-0.023) |
| Support Vector Machine | 0.658 (+/-0.032) | 0.629 (+/-0.036) | 0.669 (+/-0.048) | 0.641 (+/-0.035) | 0.72 (+/-0.032) |
| K-nearest Neighbor | 0.672 (+/-0.018) | 0.649 (+/-0.02) | 0.644 (+/-0.029) | 0.641 (+/-0.022) | 0.694 (+/-0.016) |
| Multi-layer Perceptron | 0.734 (+/-0.017) | 0.736 (+/-0.02) | 0.705 (+/-0.019) | 0.705 (+/-0.028) | 0.8 (+/-0.02) |

Table E. Performance evaluation of different models using raw data prediction results

| | accuracy | precision | recall | f1 | auc |
|---|---|---|---|---|---|
| Logistic Regression | 0.711 (+/-0.015) | 0.733 (+/-0.019) | 0.607 (+/-0.019) | 0.659 (+/-0.019) | 0.794 (+/-0.017) |
| Decision Tree | 0.728 (+/-0.022) | 0.692 (+/-0.022) | 0.758 (+/-0.035) | 0.719 (+/-0.023) | 0.779 (+/-0.021) |
| Support Vector Machine | 0.722 (+/-0.02) | 0.728 (+/-0.023) | 0.653 (+/-0.028) | 0.682 (+/-0.026) | 0.805 (+/-0.018) |
| K-nearest Neighbor | 0.709 (+/-0.015) | 0.692 (+/-0.018) | 0.694 (+/-0.023) | 0.687 (+/-0.016) | 0.776 (+/-0.022) |
| Multi-layer Perceptron | 0.735 (+/-0.015) | 0.733 (+/-0.018) | 0.689 (+/-0.019) | 0.706 (+/-0.016) | 0.809 (+/-0.014) |

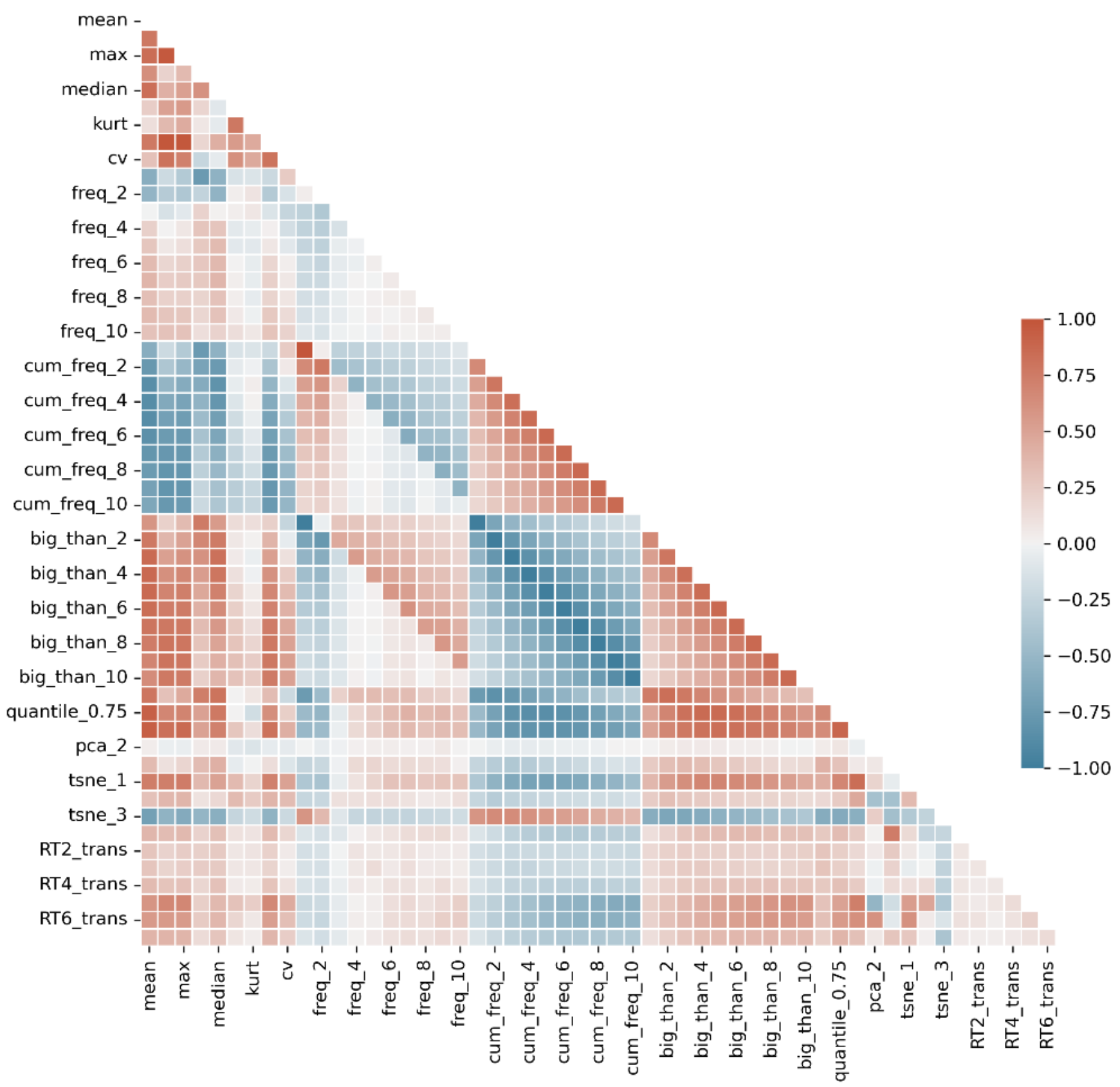

Figure A. The Correlation between Features

## Conflicts of Interest

We have no conflicts of interest to declare.

## Funding

This work was supported by National Natural Science Foundation of China [grant numbers 62176129]; National Science Fund for Distinguished Young Scholars [grant numbers 81725005]; the National Natural Science Foundation Regional Innovation and Development Joint Fund [grant numbers U20A6005]; Jiangsu Provincial Key Research and Development Program, China [grant numbers BE2021617]; Henan Provincial Research and Practice Project for Higher Education Teaching Reform [grant numbers 2021SJGLX189Y].

## Contributors

Xizhe Zhang: Conceptualization; Methodology; Supervision; Writing - Review & Editing. Fei Wang: Conceptualization; Resources; Writing - Review & Editing. Zhao Su: Data curation; Formal analysis; Methodology; Investigation; Writing - Original Draft. Xinru Wei: Methodology; Writing - Review & Editing. Keyin Zhou, Ning Wang, Zexin Lin, Yuanchen Xie, Jie Wang, Rongxun Liu, Shenzhong Zhang: Writing - Review & Editing.

## Abbreviations

RT: Response Time
ISI: Insomnia Severity Index
PCA: Principal Component Analysis
t-SNE: t-distributed Stochastic Neighbor Embedding
SVM: Support Vector Machine
KNN: K-Nearest Neighbour
MLP: Multilayer Perceptron
ROC: Receiver Operating Characteristic
SHAP: Shapley Additive Explanations